\documentclass[journal,transmag]{IEEEtran}
\usepackage{booktabs}
\usepackage{multirow}
\usepackage{float}

\usepackage{cite}

\usepackage{graphicx}
\ifCLASSINFOpdf
\else
\fi
\usepackage{amsmath}
\usepackage{amssymb}
\usepackage{caption}
\begin{document}
%
\title{Detecting Engagement in Egocentric Video}
%
%
%

\author{
\IEEEauthorblockN{Yu-Chuan~Su and
Kristen~Grauman}
\IEEEauthorblockA{Department of Computer Science, The University of Texas at Austin}
}

\maketitle

\begin{abstract}
In a wearable camera video, we see what the camera wearer sees.  
While this makes it easy to know roughly \emph{\underline{what} he chose to look at}, 
it does not immediately reveal \emph{\underline{when} he was engaged with the environment}.  
Specifically, at what moments did his focus linger, as he paused to gather more information about something he saw? 
Knowing this answer would benefit various applications in video summarization and augmented reality,  
yet prior work focuses solely on the ``what" question (estimating saliency, gaze) without considering the ``when" (engagement).
We propose a learning-based approach that uses long-term egomotion cues to detect engagement, specifically in browsing
scenarios where one frequently takes in new visual information (e.g., shopping, touring).  
We introduce a large, richly annotated dataset for ego-engagement that is the first of its kind.  Our
approach outperforms a wide array of existing methods.  
We show engagement can be detected well independent of both scene appearance and the camera wearer's identity.
\end{abstract}


\section{Introduction}

Imagine you are walking through a grocery store.   You may be mindlessly
plowing through the aisles grabbing your usual food staples, when a new product
display---or an interesting fellow shopper---captures your interest for a few
moments.  Similarly, in the museum, as you wander the exhibits, occasionally
your attention is heightened and you draw near to examine something more
closely.

These examples illustrate the notion of \emph{engagement} in ego-centric
activity, where one pauses to inspect something more closely.  While engagement
happens throughout daily life activity, it occurs frequently and markedly
during \emph{``browsing" scenarios} in which one traverses an area with the
intent of taking in new information and/or locating certain objects---for
example, in a shop, museum, library, city sightseeing, or touring a campus or
historic site.

\begin{figure}[t]
  \centering
  \begin{center}
    \includegraphics[width=1.\linewidth,clip,trim={0.4cm 0 0 0}]{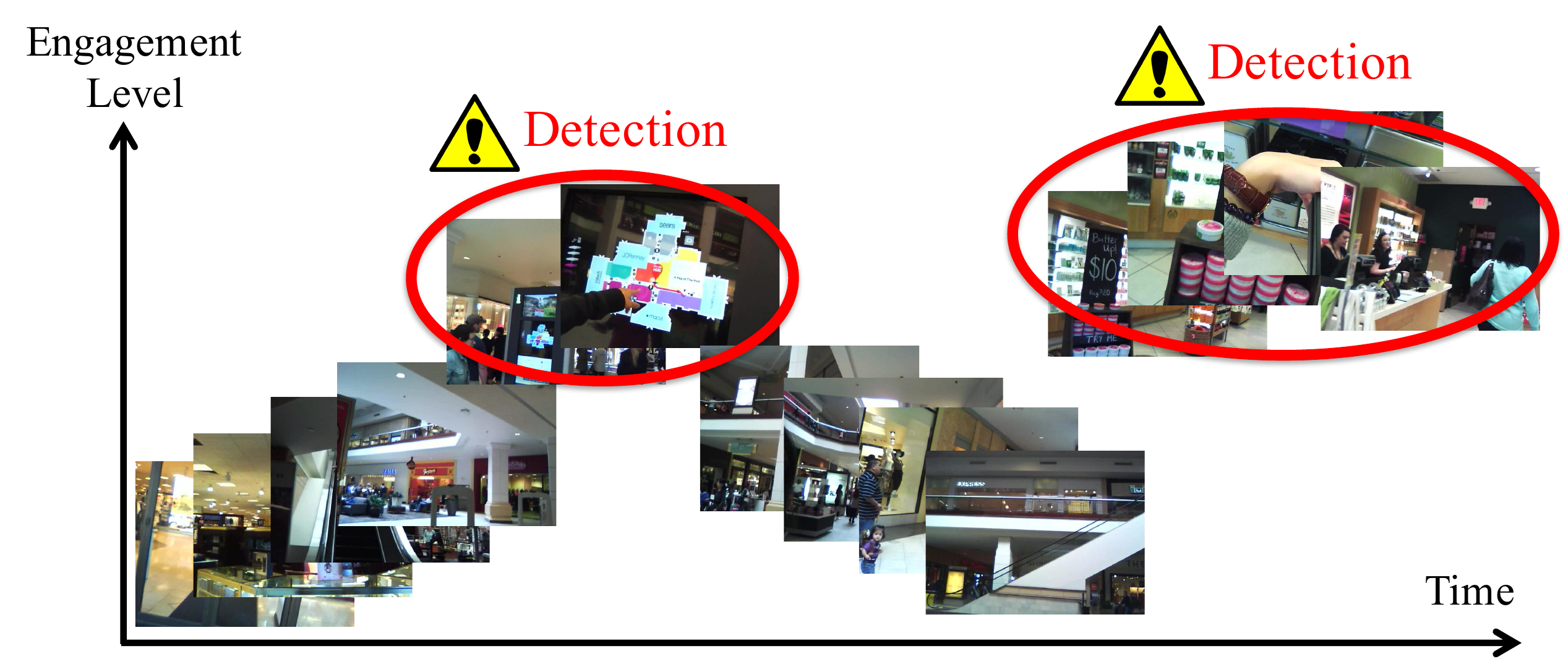}
  \end{center}
  \caption{The goal is to identify intervals where the camera wearer's
      engagement is heightened, meaning he interrupts his ongoing activity to
      gather more information about some object in the environment.  Note that
      this is different than detecting what the camera wearer sees or gazes
      upon, which comes for ``free" with a head-mounted camera and/or eye
      tracking devices.
  }
  \label{fig:concept}
\end{figure}

\paragraph{Problem definition}
We explore engagement  from the first-person vision perspective.  In
particular, we ask: Given a video stream captured from a head-mounted camera
during a browsing scenario, can we automatically detect those time intervals
where the recorder experienced a heightened level of engagement?  What cues are
indicative of first-person engagement, and how do they differ from traditional
saliency metrics?  To what extent are engagement cues independent of the
particular person wearing the camera (the ``recorder''), or the particular
environment they are navigating?  See Fig.~\ref{fig:concept}.

While engagement is interesting in a variety of daily life settings, for now we
restrict our focus to browsing scenarios.  This allows us to concentrate on
cases where 1) engagement naturally ebbs and flows repeatedly, 2) the
environment offers discrete entities (products in the shop, museum paintings,
etc.) that may be attention-provoking, which aids objectivity in evaluation,
and 3) there is high potential impact for emerging applications.

\paragraph{Applications}
A system that can successfully address the above questions would open up
several applications.  For example, it could facilitate camera control,
allowing the user's attention to trigger automatic recording/zooming.
Similarly, it would help construct video summaries.  Knowing when a user's
engagement is waning would let a system display info on a heads-up display when
it is least intrusive.  Beyond such ``user-centric'' applications, third
parties would relish the chance to gather data about user attention at
scale---for instance, a vendor would like to know when shoppers linger by its
new display. Such applications are gaining urgency as wearable cameras become
increasingly attractive tools in the law enforcement, healthcare, education,
and consumer domains.

\paragraph{Novelty of the problem}
The rich literature on visual saliency---including video
saliency~\cite{rudoy2013cvpr,han2014neurocomputing,lee-tip2011,abdollahian2010tmm,mahadevan-pami2010,rahtu-eccv2010,itti-2009,liu2008mm}---does
not address this problem.  First and foremost, as discussed above, detecting
moments of engagement is different than estimating saliency.  Nearly all prior
work studies visual saliency from the \emph{third person} perspective and
equates saliency with gaze: salient points are those upon which a viewer would
fixate his gaze, when observing a previously recorded image/video on a static
screen.  In contrast, our problem entails detecting \emph{temporal intervals of
engagement as perceived by the person capturing the video as he moves about his
environment}.  Thus, \emph{recorder engagement} is distinct from \emph{viewer
attention}.  To predict it from video requires identifying time intervals of
engagement as opposed to spatial regions that are salient (gaze worthy) per
frame.  As such, estimating egocentric
gaze~\cite{li2013iccv,yamada2011psivt,yamada2010accv} is also insufficient to
predict first-person engagement.

\paragraph{Challenges}
Predicting first-person engagement presents a number of challenges.  First of
all, the motion cues that are significant in third-person video taken with an
actively controlled camera (e.g.,
zoom~\cite{kender2000accv,li2012bmvc,gygli-eccv2014,abdollahian2010tmm}) are
absent in passive wearable camera data.  Instead, first-person data contains
both scene motion and unstable body motions, which are difficult to stabilize
with traditional methods~\cite{poleg2014cvpr}.  Secondly, whereas third-person
data is inherently already focused on moments of interest that led the recorder
to turn the camera on, a first-person camera is ``always on".  Thirdly, whereas
traditional visual attention metrics operate with instantaneous motion
cues~\cite{rudoy2013cvpr,nguyen2013mm,ejaz-2013,han2014neurocomputing} and
fixed sliding temporal window search strategies, detecting engagement
\emph{intervals} requires long-term descriptors and handling intervals of
variable length.  Finally, it is unclear whether there are sufficient visual
cues that transcend user- or scene-specific properties, or if engagement is
strongly linked to the specific content a user observes (in which case, an
exorbitant amount of data might be necessary to learn a general-purpose
detector).

\paragraph{Our approach}
We propose a learning approach to detect time intervals where first-person
engagement occurs.  In an effort to maintain independence of the camera wearer
as well as the details of his environment, we employ motion-based features that
span long temporal neighborhoods and integrate out local head motion effects.
We develop a search strategy that integrates instantaneous frame-level
estimates with temporal interval hypotheses to detect intervals of varying
lengths, thereby avoiding a naive sliding window search.  To train and evaluate
our model, we undertake a large-scale data collection effort.

\paragraph{Contributions}
Our main contributions are as follows. First, we precisely define egocentric
engagement and systematically evaluate under that definition.  Second, we
collect a large annotated dataset spanning 14 hours of activity explicitly
designed for ego-engagement in browsing situations.  Third, we propose a
learned motion-based model for detecting first-person engagement.  Our model
shows better accuracy than an array of existing methods.  It also generalizes
to unseen browsing scenarios, suggesting that some properties of ego-engagement
are independent of appearance content.

\section{Related Work}
\subsection{Third-person image and video saliency}

Researchers often equate human gaze fixations as the gold standard with which a
\emph{saliency} metric ought to correlate~\cite{itti-motion2003,harel2006nips}.
There is increasing interest in estimating saliency from video.  Initial
efforts examine simple motion cues, such as frame-based motion and
flicker~\cite{itti-motion2003,liu2008mm,seo-2009}.  One common approach to
extend spatial (image) saliency to the video domain is to sum image saliency
scores within a temporal segment, e.g.,~\cite{ma2002mm}.  Most methods are
unsupervised and entail no
learning~\cite{itti-motion2003,itti-2009,liu2008mm,mahadevan-pami2010,abdollahian2010tmm,rahtu-eccv2010,seo-2009}.
However, some recent work  develops learned measures,  using ground truth gaze
data as the target
output~\cite{kienzle-dagm2007,lee-tip2011,rudoy2013cvpr,nguyen2013mm,han2014neurocomputing}.

Our problem setting is quite different than saliency.  Saliency aims to
\emph{predict viewer attention} in terms of where in the frame a third party is
likely to fixate his gaze; it is an image property analyzed independent of the
behavior of the person recording the image.  In contrast, we aim to
\emph{detect recorder engagement} in terms of when (which time intervals) the
recorder has paused to examine something in his
environment.\footnote{Throughout, we will use the term ``recorder" to refer to
the
photographer or the first-person camera-wearer; we use the term ``viewer" to
refer to a third party who is observing the data captured by some other
recorder.}    Accounting for
this distinction is crucial, as we will see in results.  Furthermore, prior
work in video saliency is evaluated on short video clips (e.g., on the order of
10 seconds~\cite{dorr2010jvis}), which is sufficient to study gaze movements.
In contrast, we evaluate on long sequences---30 minutes on average per clip,
and a total of 14 hours---in order to capture the broad context of ego-behavior
that affects engagement in browsing scenarios.

\subsection{Third-person video summarization}
In video summarization, the goal is to form a concise representation for a long
input video.  Motion cues can help detect ``important" moments in third-person
video~\cite{kender2000accv,ma2002mm,li2012bmvc,ejaz-2013,gygli-eccv2014},
including temporal differences~\cite{ejaz-2013} and cues from active camera
control~\cite{kender2000accv,li2012bmvc,gygli-eccv2014}.  Whereas prior methods
try to extract what will be interesting to a third-party viewer, we aim to
capture \emph{recorder} engagement.

\subsection{First-person video saliency and gaze}

Researchers have long expected that ego-attention detection requires methods
distinct from bottom-up saliency~\cite{hp}.  In fact, traditional motion
saliency can actually \emph{degrade} gaze prediction for first-person
video~\cite{yamada2010accv}.  Instead, it is valuable to separate out camera
motion~\cite{yamada2011psivt} or use head motion and hand locations to predict
gaze~\cite{li2013iccv}.  Whereas these methods aim to predict spatial
coordinates of a recorder's gaze at every frame, we aim to predict time
intervals where his engagement is heightened.  Furthermore, whereas they study
short sequences in a lab~\cite{yamada2011psivt} or kitchen~\cite{li2013iccv},
we analyze long data in natural environments with substantial scene changes per
sequence.  

We agree that first-person attention, construed in the most general sense, will
inevitably require first-person ``user-in-the-loop" feedback to
detect~\cite{hp}; accordingly, our work does not aim to detect arbitrary
subjective attention events, but instead to detect  moments of engagement to
examine an object more closely.

Outside of gaze, there is limited work on attention in terms of head fixation
detection~\cite{poleg2014cvpr} and ``physical
analytics"~\cite{rallapalli-mobicom2014}.  In~\cite{poleg2014cvpr}, a novel
``cumulative displacement curve" motion cue is used to categorize the
recorder's activity (walking, sitting, on bus, etc.) and is also shown to
reveal periods with fixed head position.  They use a limited definition of
attention: a period of more than 5 seconds where the head is still but the
recorder is walking.  In~\cite{rallapalli-mobicom2014}, inertial sensors are
used in concert with optical flow magnitude to decide when the recorder is
examining a product in a store.  Compared to
both~\cite{rallapalli-mobicom2014,poleg2014cvpr}, engagement has a broader
definition, and we discover its scope from data from the crowd
(vs.~hand-crafting a definition on visual features).  Crucially, the true
positives reflect that a person can have heightened engagement yet still be in
motion.

\subsection{First-person activity and summarization}

Early methods for egocentric video summarization extract the camera motion and
define rules for important moments (e.g., intervals when camera rotation is
below a threshold)~\cite{nakamura2000pr,cheatle2004icpr}, and test
qualitatively on short videos.  Rather than inject hand-crafted rules, we
propose to \emph{learn} what constitutes an engagement interval.  Recent
methods explore ways to predict the ``importance" of spatial regions (objects,
people) using cues like hand detection and frame
centrality~\cite{lee-cvpr2012,lu-cvpr2013}, detect
novelty~\cite{carlsson-cvpr2011}, and infer ``social saliency" when multiple
cameras capture the same
event~\cite{hoshen2014cvpr,park-nips2012,fathi-cvpr2012}.  We tackle
engagement, not summarization, though likely our predictions could be another
useful input to a summarization system.

In a sense, detecting engagement could be seen as detecting a particular
ego-activity.  An array of methods for classifying activity in egocentric video
exist,
e.g.,~\cite{fathi,peleg,pirsiavash,damen,farhadi,kitani-activity,spriggs,yinli}.
However,  they do not address our scenario: 1) they learn models specific to
the objects~\cite{fathi,yinli,pirsiavash,damen,spriggs,farhadi} or
scenes~\cite{kitani-activity} with which the activity takes place (e.g., making
tea, snowboarding), whereas engagement is by definition object- and
scene-independent, since arbitrary things may capture one's interest; and 2)
they typically focus on recognition of trimmed video clips, versus temporal
detection in ongoing video.

\section{First-Person Engagement: Definition and Data}

Next we define first-person engagement.  Then we describe our data collection procedure, and quantitatively
analyze the consistency of the resulting annotations.  We introduce our
approach for predicting engagement intervals in Sec.~\ref{sec:approach}.

\subsection{Definition of first-person engagement}
\label{sec:definition}

This research direction depends crucially on having (1) a precise definition of
engagement, (2) realistic video data captured in natural
environments, and (3) a systematic way to annotate the data for both learning
and evaluation.

Accordingly, we first formalize our meaning of first-person engagement. There
are two major requirements. First, the engagement must be related to
external factors, either induced by or causing the change in visual signals
the recorder perceives. This ensures predictability from video, excluding
high-attention events that are imperceptible (by humans) from  visual cues.
Second, an engagement interval must reflect the \emph{recorder's} intention,
as opposed to the reaction of a third-person viewer of the same video.

Based on these requirements, we \textbf{define heightened ego-engagement in a
browsing scenario} as follows. A time interval is considered to have a high
engagement level if \emph{the recorder is attracted by some object(s), and he
    interrupts his ongoing flow of activity to purposefully gather more
information about the object(s).}  We stress that this definition is scoped
specifically for \emph{browsing} scenarios; while the particular objects
attracting the recorder will vary widely, we assume the person is traversing
some area with the intent of taking in new information and/or locating certain
objects.

The definition captures situations where the recorder reaches out to touch or
grasp an object of interest (e.g., when closely inspecting a product at the
store), as well as scenarios where he examines something from afar (e.g., when
he reads a sign beside a painting at the museum).  Having an explicit
definition allows annotators to consistently identify video clips with high
engagement, and it lets us directly evaluate the prediction result of different
models.

We stress that ego-engagement differs from gaze and traditional saliency.
While a recorder always has a gaze point per frame (and it is correlated with
the frame center), periods of engagement are more sparsely distributed across
time, occupy variable-length intervals, and are a function of his activity and
changing environment.  Furthermore, as we will see below, moments where a
person is approximately still are \emph{not} equivalent to moments of
engagement, making observer motion magnitude~\cite{rallapalli-mobicom2014} an
inadequate signal.

\subsection{Data collection}
\label{sec:datacollection}

To collect a dataset, 
we ask multiple recorders to take videos during ``browsing'' behavior under a set of \emph{scenarios},
or scene and event types.
We aim to gather scenarios with clear distinctions between high and low engagement intervals
that will be apparent to a third-party annotator.
Based on that criterion,
we collect videos under three scenarios:
(1) shopping in a market,
(2) window shopping in shopping mall, and
(3) touring in a museum.
All three entail spontaneous stimuli, which ensures that variable levels of engagement will naturally occur.

The videos are recorded using Looxcie LX2 with $640 \times 480$ resolution
and 15 fps frame rate, which we chose for its long battery life and low profile. We recruited 9 recorders---5 females and 4 males---all
students between 20-30 years old.
Other than asking them to capture
instances of the scenarios above, we did not otherwise instruct the recorders
to behave in any way.
Among the 9 recorders, 5 of them record videos in all
3 scenarios. The other 4 record videos in 2 scenarios. Altogether, we
obtained 27 videos, each averaging 31 minutes, for a total dataset
of 14 hours.
To keep the recorder behavior as natural as possible,
we asked the recorders to capture the video when they planned to go to such scenarios anyway;
as such, it took about 1.5 months to collect the video.

After collecting the videos, we crowdsource the ground truth annotations on
Amazon Mechanical Turk.  Importantly, we ask annotators to put themselves in
the camera-wearer's shoes.  They must precisely mark the start and end points
of each engagement interval from the recorder's perspective, and record their
confidence.\footnote{For a portion of the video, we also ask the original
recorders to label all frames for their own video; this requires substantial
tedious effort, hence to get the full labeled set in a scalable manner we apply
crowdsourcing.} We break the source videos into 3 minutes overlapping chunks to
make each annotation task manageable yet still reveal temporal context for the
clip.  We estimate the annotations took about 450 worker-hours and cost
\$3,000.  Our collection strategy is congruous with the goals stated above in
Sec.~\ref{sec:definition}, in that annotators are shown only the visual signal
(without audio) and are asked to consider engagement from the point of view of
the recorder.  See appendix for the details of annotation process.

\begin{table}[t]\small
    \begin{center}
    \begin{tabular}{lcccc}
        \toprule
                                    & Mall  & Market& Museum& All\\
        \midrule
        Attention Ratio         & 0.305 & 0.451 & 0.580 & 0.438\\
        \#intervals (per min.)      & 1.19  & 1.22  & 1.50  & 1.30 \\
        Length Median (sec)     & 7.5   & 12.1  & 13.3  & 11.3 \\
        Length IQR (sec)    & 11.6  & 18.2  & 20.1  & 17.6 \\
        \bottomrule
    \end{tabular}
    \caption{Basic statistics for ground truth intervals.\label{table:stats}}
    \end{center}
\end{table}

Despite our care in the instructions, there remains room for annotator
subjectivity, and the exact interval boundaries can be ambiguous.  Thus, we ask
10 Turkers to annotate each video.  Positive intervals are those where a
majority agree engagement is heightened.  To avoid over-segmentation, we ignore
intervals shorter than 1 second.  For each positive interval, we select the
tightest annotation that covers more than half of the interval as the final
ground truth.

The resulting dataset contains examples that are diverse in content and
duration.  The recorders are attracted by a variety of objects: groceries,
household items, clothes, paintings, sculptures, other people.  In some cases,
the attended object is out of the field of view, e.g., a recorder grabs an item
without directly looking at it, in which case Turkers infer the engagement from
context.

Table~\ref{table:stats} summarizes some statistics of the labeled data.  On
average, the recorder is engaged about 44\% of the time (see ``Attention
Ratio''), and it increases once to twice per minute.  This density reflects the
``browsing'' scenarios on which we focus the data.  The length of a positive
interval varies substantially: the interquartile range (IQR) is 17.6 seconds,
about 50\% longer than the median.  Some intervals last as long as 5 minutes.
Also, different scenarios have different statistics, e.g., Museum scenarios
prompt more frequent engagement.  All this variability indicates the difficulty
of the task.

The new dataset is the first of its kind to explicitly define and thoroughly
annotate ego-engagement.  It is also substantially larger than datasets used in
related areas---nearly 14 hours of video, with test videos over 30 minutes
each.  By contrast, clips in popular third-person saliency datasets are
typically 20 seconds~\cite{dorr2010jvis} to 2 minutes~\cite{mital2011cogcomp},
since the interest is in gauging instantaneous gaze reactions.

\subsection{Evaluating data consistency}

How consistently do third-party annotators label
engagement intervals?  We analyze their consistency to verify the
predictability and soundness of our definition.

Table~\ref{tab:annotationanalysis} shows the analysis.  We quantify label
agreement in terms of the average $F_{1}$ score, whether at the frame or
interval level (see Sec.~\ref{sec:evaluation_metric}).  We consider two aspects
of agreement: boundary (how well do annotators agree on the start and end
points of a positive interval?) and presence (how well do they agree on the
existence of a positive interval?).

First we compare how consistent each of the 10 annotators' labels are with the
consensus ground truth (see ``Turker vs.~Consensus").  They have reasonable
agreement on the rough interval locations, which verifies the soundness of our
definition. Still, the $F_{1}$ score is not perfect, which indicates that the
task is non-trivial even for humans.  Some discrepancies are due to the fact
that even when two annotators agree on the presence of an interval, their
annotations will not match exactly in terms of the start and end frame.  For
example, one annotator might mark the start when the recorder searches for
items on the shelf, while another might consider it to be when the recorder
grabs the item.  Indeed, agreement on the presence criterion (right column) is
even higher, 0.914.  The ``Random vs.~Consensus'' entry compares a
prior-informed random guess to the ground truth.\footnote{We randomly generate
    interval predictions 10 times based on the prior of interval length and
temporal distribution and report the average.} These two extremes give useful
bounds of what we can expect from our computational model: a predictor should
perform better than random, but will not exceed the inter-human agreement.

\begin{table}[t]\small
  \begin{center}
  \begin{tabular}{rclccc}
      \toprule
      &&& \multirow{2}{*}{Frame $F_{1}$} & \multicolumn{2}{c}{Interval $F_{1}$}\\
      \cmidrule{5-6}
      &&&                                & Boundary & Presence\\
      \midrule
      \multirow{2}*{Turker} &vs.~& Consensus & 0.818 & 0.837 & 0.914\\
                            &vs.~& Recorder & 0.589 & 0.626 & 0.813 \\
      \midrule
      \multirow{2}{*}{Random} &vs.~& Consensus & 0.426 & 0.339 & 0.481 \\
                              &vs.~& Recorder & 0.399 & 0.344 & 0.478 \\
      \bottomrule
  \end{tabular}
  \caption{Analysis of inter-annotator consistency.}
  \label{tab:annotationanalysis}
  \end{center}
\end{table}

Next, we check how well the third-party labels match the experience of the
first-person recorder (see ``Turker vs.~Recorder).  We collect 3 hours of
self-annotation from 4 of the recorders, and compare them to the Turker
annotations.  Similar to above, we see the Turkers are considerably more
consistent with the recorder labels compared to the prior-informed random
guess, though not perfect.  As one might expect, Turker annotations have higher
recall, but lower (yet reasonable) precision against the first-person labels.
Overall, the 0.813 $F_1$ score for Turker-Recorder presence agreement indicates
our labels are fairly faithful to individuals' subjective interpretation.

\section{Approach}
\label{sec:approach}
We propose to learn the motion patterns in first-person video that indicate
engagement.  Two key factors motivate our decision to focus on motion.  First,
camera motion often contains useful information about the recorder's
intention~\cite{kender2000accv,li2012bmvc,yamada2011psivt}.  This is especially
true in egocentric video, where the recorder's head and body motion heavily
influence the observed motion.  Second, motion patterns stand to generalize
better across different scenarios, as they are mostly independent of the
appearance of the surrounding objects and scene.

\begin{figure}[t]
  \centering
  \includegraphics[width=1.\linewidth,clip,trim={0 0.2cm 0 0}]{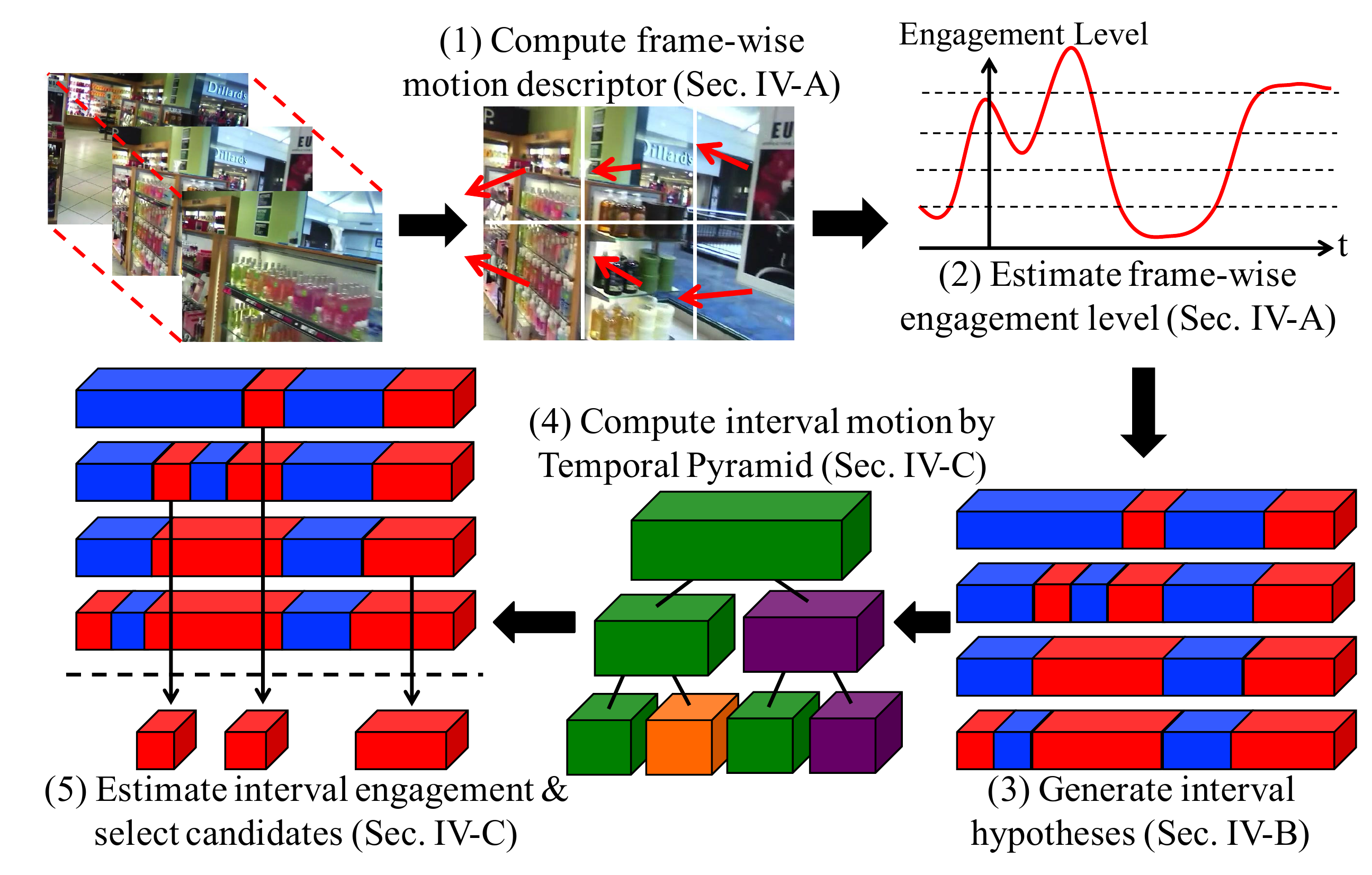}
  \caption{Workflow for our approach.}
  \label{fig:approach}
\end{figure}

Our approach has three main stages.  First we compute \emph{frame-wise}
predictions (Sec.~\ref{sec:framewise_estimation}).  Then we leverage those
frame predictions to generate \emph{interval} hypotheses
(Sec.~\ref{sec:candidategeneration}).  Finally, we describe each interval as a
whole and classify it with an interval-trained model (Sec.~\ref{sec:interval}).
By departing from traditional frame-based
decisions~\cite{nakamura2000pr,cheatle2004icpr,ejaz-2013}, we capture long-term
temporal dependencies.  As we will see below, doing so is beneficial for
detecting subtle periods of engagement and accounting for their variable
length.  Fig.~\ref{fig:approach} shows the workflow.

\subsection{Initial frame-wise estimates}
\label{sec:framewise_estimation}

To first compute frame-wise predictions, we construct one motion descriptor per
frame.  We divide the frame into a grid of $16 \times 12$ uniform cells and
compute the optical flow vector in each cell.  Then we temporally smooth the
grid motion with a Gaussian kernel.  Since at this stage we want to capture
attention within a granularity of a second, we set the width of the kernel to
two seconds.  As shown in~\cite{poleg2014cvpr}, smoothing the flow is valuable
to integrate out the regular unstable head bobbles by the recorder; it helps
the descriptor focus on prominent scene and camera motion.  The frame
descriptor consists of the smoothed flow vectors concatenated across cells,
together with the mean and standard deviation of all cells in the frame.  It
captures dominant egomotion and dynamic scene motion---both of which are
relevant to first-person engagement.

We use these descriptors, together with the frame-level ground truth
(cf.~Sec.~\ref{sec:datacollection}), to train an i.i.d. classifier.  We use
random forest classifiers due to their test-time efficiency and relative
insensitivity to hyper-parameters, though of course other classifiers are
possible.  Given a test video, the confidence (posterior) output by the random
forest is used as the initial frame-wise engagement estimate.

\subsection{Generating interval proposals}
\label{sec:candidategeneration}

After obtaining the preliminary estimate for each frame, we generate multiple
hypotheses for engagement \emph{intervals} using a level set method as follows.
For a given threshold on the frame-based confidence, we obtain a set of
positive intervals, where each positive interval consists of contiguous frames
whose confidence exceeds the threshold.  By sweeping through all possible
thresholds (we use the decile), we generate multiple such sets of candidates.
Candidates from all thresholds are pooled together to form a final set of
\emph{interval proposals}.

We apply this candidate generation process on both training data and test data.
During training, it yields both positive and negative example intervals that we
use to train an interval-level classifier (described next).  During testing, it
yields the hypotheses to which the classifier should be applied.  This
detection paradigm not only lets us avoid sliding temporal window search, but
it also allows us to detect engagement intervals of variable length.

\subsection{Describing and classifying intervals}\label{sec:interval}

For each interval proposal, we generate a motion descriptor that captures both
the motion distribution and evolution over time.  Motion evolution is important
because a recorder usually performs multiple actions within an interval of
engagement.  For example, the recorder may stop, turn his head to stare at an
object, reach out to touch it, then turn back to resume walking.  Each action
leads to a different motion pattern.  Thus, unlike the temporally local
frame-based descriptor above, here we aim to capture the statistics of the
entire interval. We'd also like the representation to be robust to time-scale
variations (i.e., yielding similar descriptors for long and short instances of
the same activity).

To this end, we use a temporal pyramid representation.  For each level of the
pyramid, we divide the interval from the previous level into two equal-length
sub-intervals.  For each sub-interval, we aggregate the frame motion computed
in Sec.~\ref{sec:framewise_estimation} by taking the dimension-wise mean and
variance.  So, the top level aggregates the motion of the entire interval, and
its descendants aggregate increasingly finer time-scale intervals.  The
aggregated motion descriptors from all sub-intervals are concatenated to form a
temporal pyramid descriptor.  We use 3-level pyramids.  To provide further
context, we augment this descriptor with those of its temporal neighbor
intervals (i.e., before and after).  This captures the motion \emph{change}
from low engagement to high engagement and back.

We train a random forest classifier using this descriptor and the interval
proposals from the training data, this time referring to the interval-level
ground truth from Sec.~\ref{sec:datacollection}.  At test time, we apply this
classifier to a test video's interval proposals to score each one.  If a frame
is covered by multiple interval proposals, we take the highest confidence score
as the final prediction per frame.

\subsection{Discussion}

Our method design is distinct from previous work in video \emph{attention},
which typically operates per frame and uses temporally local measurements of
motion~\cite{nakamura2000pr,cheatle2004icpr,rudoy2013cvpr,nguyen2013mm,ejaz-2013,han2014neurocomputing}.
In contrast, we estimate enagement from interval hypotheses bootstrapped from
initial frame estimates, and our representation captures motion changes over
time at multiple scales. People often perform multiple actions during an
engagement interval, which is well-captured by considering an interval
together.  For example, it is hard to tell whether the recorder is attracted by
an object when we only know he glances at it, but it becomes clear if we know
his following action is to turn to the object or to turn away quickly.

Simply flagging periods of low
motion~\cite{rallapalli-mobicom2014,poleg2014cvpr,cheatle2004icpr} is
insufficient to detect all cases of heightened attention, since behaviors
during the interval of engagement are often non-static and also exhibit
learnable patterns.  For example, shoppers move and handle objects they might
buy; people sway while inspecting a painting; they look up and sweep their gaze
downward when inspecting a skyscraper.

External sensors beyond the video stream could potentially provide cues useful
to our task, such as inertial sensors to detect recorder motion and head
orientation.  However, such sensors are not always available, and they are
quite noisy in practice.  In fact, recent attempts to detect gazing behavior
with inertial sensors alone yield false positive rates of
33\%~\cite{rallapalli-mobicom2014}.  This argues for the need for visual
features for the challenging engagement detection task.

\section{Experiments}
\subsection{Experiment Setting}
We validate on two datasets and compare to many existing methods.

\paragraph{Baselines}

We compare with 9 existing methods, organized into four types:
\begin{itemize}
\item \textbf{Saliency Map}: Following \cite{ma2002mm,ejaz-2013}, we compute
    the saliency map for each frame and take the average saliency value.  We
    apply the state-of-the-art learned video saliency model
    \cite{rudoy2013cvpr} and five others that were previously used for video
    summarization:
    \cite{seo-2009,harel2006nips,itti-2009,ejaz-2013,rahtu-eccv2010}. We use
    the original authors' code
    for~\cite{seo-2009,harel2006nips,itti-2009,rahtu-eccv2010,rudoy2013cvpr}
    and implement \cite{ejaz-2013}. Except \cite{rahtu-eccv2010}, all these
    models use motion.
    
\item \textbf{Motion Magnitude}: Following~\cite{cheatle2004icpr,rallapalli-mobicom2014}, this baseline uses
    the inverse motion magnitude.  Intuitively, the recorder becomes more still
    during his moments of  high engagement as he inspects the object(s).  We
    apply the same flow smoothing as in Sec.~\ref{sec:framewise_estimation} and
    take the average.

\item \textbf{Learned Appearance (CNN)}: This baseline predicts engagement
    based on the video content.  We use state-of-the-art convolutional neural
    net (CNN) image descriptors, and train a random forest with the same
    frame-based ground truth our method uses.  We use Caffe~\cite{jia2014caffe}
    and the provided pre-trained model (BVLC Reference CaffeNet).

\item \textbf{Egocentric Important Region}: This is the method
    of~\cite{lee-cvpr2012}.  It is a learned metric designed for egocentric
    video that exploits hand detection, centrality in frame, etc. to predict
    the importance of regions for summarization.  While the objective of
    ``importance" is different than ``engagement", it is related and valuable
    as a comparison, particularly since it also targets egocentric data. We
    take the max importance per frame using the predictions shared by the
    authors. 
\end{itemize}

Some of the baselines do not target our task specifically, a likely
disadvantage.  Nonetheless, their inclusion is useful to see if ego-engagement
requires methods beyond existing saliency metrics.  Besides, our baselines also
include methods specialized for egocentric
video~\cite{rallapalli-mobicom2014,lee-cvpr2012}, and one that targets exactly
our task~\cite{rallapalli-mobicom2014}.

For the learned methods (ours, CNN, and Important Regions), we use the
classifier confidences to rate frames by their engagement level.   Note that
the CNN method has the benefit of training on the exact same data as our
method.  For the non-learned methods (saliency, motion), we use their
magnitude.  We evaluate two versions of our method: one with the interval
proposals (Ours-interval), and one without (Ours-frame).  The boundary
agreement is used for interval prediction evaluation to favor methods with
better localization of engagement.

\paragraph{Datasets}

We evaluate on two datasets: our new UT Egocentric Engagement (UT EE) dataset and the
public UT Egocentric dataset (UT Ego).  We select all clips from UT Ego that
contain browsing scenarios (mall, market), yielding 3 clips with total length
of 58 minutes, and get them annotated with the same procedure in
Sec.~\ref{sec:datacollection}.  

\begin{table}[t]\scriptsize
    \tabcolsep=0.1cm
    \begin{center}
    \begin{tabular}{llcc}
    \toprule
    & & Frame $F_{1}$ & Interval $F_{1}$\\
    \midrule
    GBVS & (Harel~2006~\cite{harel2006nips})                      & 0.462 & 0.286 \\
    Self Resemblance & (Seo~2009~\cite{seo-2009})                 & 0.471 & 0.398 \\
    Bayesian Surprise & (Itti~2009~\cite{itti-2009} )             & 0.420 & 0.373 \\
    Salient Object & (Rahtu~2010~\cite{rahtu-eccv2010})           & 0.504 & 0.389 \\
    Video Attention & (Ejaz~2013~\cite{ejaz-2013})                & 0.413 & 0.298 \\
    Video Saliency & (Rudoy~2013~\cite{rudoy2013cvpr})            & 0.435 & 0.396 \\
    Motion Mag. & (Rallapalli~2014~\cite{rallapalli-mobicom2014}) & 0.553 & 0.403 \\
    \midrule
    \multirow{4}{*}{Cross Recorder} & CNN Appearance            & 0.685 & 0.486 \\
    & Ours -- frame                                             & \textbf{0.686} & 0.533 \\
    & Ours -- interval                                          & 0.674 & \textbf{0.572} \\
    \cmidrule{2-4}
    & Ours -- GT interval                                       & 0.822 & 0.868 \\
    \midrule
    \multirow{4}{*}{Cross Scenario} & CNN Appearance            & 0.656 & 0.463 \\
    & Ours -- frame                                             & \textbf{0.683} & 0.531 \\
    & Ours -- interval                                          & 0.665 & \textbf{0.553} \\
    \cmidrule{2-4}
    & Ours -- GT interval                                       & 0.830 & 0.860 \\
\midrule
     \multirow{4}{*}{Cross Recorder AND Scenario} & CNN Appearance            & 0.655 & 0.463 \\
    & Ours -- frame                                             & \textbf{0.680} & 0.532 \\
    & Ours -- interval                                          & 0.661 & \textbf{0.544} \\
    \cmidrule{2-4}
    & Ours -- GT interval    & 0.823 & 0.856\\
    \bottomrule
    \end{tabular}
    \caption{$F_{1}$-score accuracy of all methods on UT EE.
        (The cross-recorder/scenario distinctions are not relevant to the top
        block of methods, all of which do no learning.)
    }
    \label{tab:temporalattention_result}
    \end{center}
\end{table}

\begin{figure*}[t]
    \begin{center}
    \makebox[\textwidth][c]{\includegraphics[width=1.0\textwidth,clip,trim={0 8.2cm 0 0}]{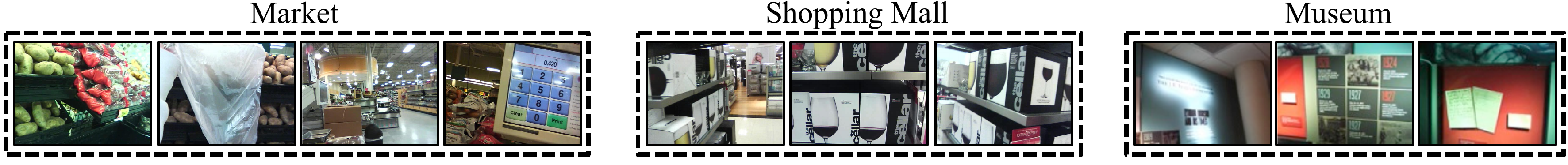}}
    \caption{Example engagement intervals detected by our method.  
        Note the intra-interval variation: the recorder either performs
        multiple actions (Market), looks at an item from multiple views (Mall)
        or looks at multiple items (Museum).
    }
    \label{fig:qual}
    \end{center}
\end{figure*}

\paragraph{Implementation details}

We use the code of \cite{liu2009phdthesis} for optical flow computation.  Flow
dominates our run-time, about 1.2 s per frame on 48 cores.  The default
settings are used for this and all the public saliency map codes.  Using the
scikit-learn package \cite{scikit-learn} for random forest, we train 2,400
trees in all results and leave all other parameters at default.  The sample
rate of video frames is 15 fps for optical flow and 1 fps for all other
computation, including evaluation.

\paragraph{Evaluation metric}
\label{sec:evaluation_metric}
We evaluate the performance of different methods using the metric defined
below.
Let $G$ denote a set of ground truth intervals for engagement.
The set of intervals is consistent if none of the intervals within the set overlap with others,
denoted by $|g_{1} \cap g_{2}|=0, \: \forall g_{1} g_{2} \in G$. 
$g_{1} \cap g_{2}$ denotes the frames that are in both interval $g_{1}$ and $g_{2}$.
Also, let $P$ denote a set of predicted intervals that is consistent.

We consider a predicted interval $p$ to be covered by a ground truth interval $g$ if
$\frac{1}{2} |p \cap g| > |p|$, denoted by $p \subset g$. 
Given the ground truth intervals $G$ and predictions $P$,
we define the interval precision as follows:
\[
	Precision = \frac{|\{\exists g \in G\;s.t.\;p \subset g \; | \; \forall p \in P\}|}{|P|}.
\]
Similarly,
we consider a ground truth interval $g$ to be covered by a predicted interval $p$ if 
$\frac{1}{2} |p \cap g| > |g|$, and we compute the interval recall as
\[
	Recall = \frac{|\{\exists p \in P\;s.t.\;g \subset p \; | \; \forall g \in G\}|}{|G|}.
\]
Note the recall monotonically increases as we prolong the length of each
prediction $p$ in $P$.  Roughly speaking, a predicted interval $p$ is
considered correct if more than $50\%$ of the prediction overlaps with some
ground truth interval, and a ground truth interval is considered predicted if
more than $50\%$ of the interval is covered by some prediction.

\subsection{UT Egocentric Engagement (UT EE) dataset}

We consider three strategies to form train-test data splits.  The first is
leave-one-recorder-out, denoted \textbf{cross-recorder}, in which we train a
predictor for each recorder using exclusively video from \emph{other}
recorders.  This setting tests the ability to generalize to new recorders
(e.g., can we learn from John's video to predict engagement in Mary's video?).
The second is leave-one-scenario-out, denoted as \textbf{cross-scenario}, in
which we train a predictor for each scenario using exclusively video from other
scenarios.  This setting examines to what extent visual cues of engagement are
independent of the specific activity or location the recorder (e.g., can we
learn from a museum trip to predict engagement during a shopping trip?).  The
third strategy is the most stringent, disallowing any overlap in either the
recorder or the scenario (\textbf{cross recorder AND scenario}).  

\begin{figure*}[t]
    \makebox[\textwidth][c]{\includegraphics[width=1.0\textwidth]{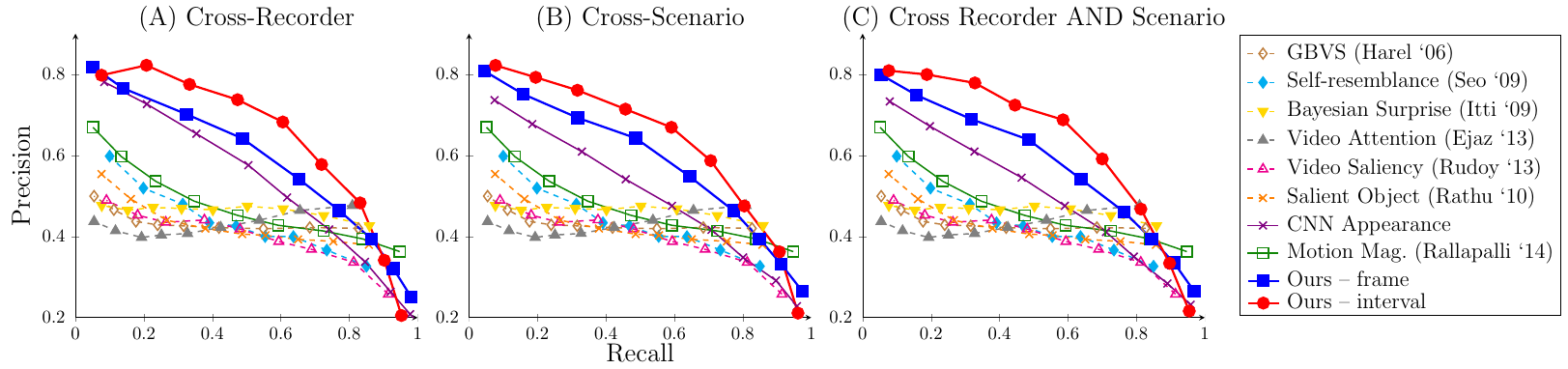}}
   \caption{Precision-recall accuracy on the UT EE dataset. Our approach detects
       engagement with much greater accuracy than an array of saliency and
       content-based methods, and our interval proposal idea improves the
       initial frame-wise predictions. 
   }
    \label{fig:temporalattention_result}
\end{figure*}

Fig.~\ref{fig:temporalattention_result}(A)$\sim$(C) show the precision-recall
curves for all methods and settings on the 14 hour UT EE dataset, and we
summarize them in Table~\ref{tab:temporalattention_result} using the
$F_{1}$ scores; here we set the confidence threshold for each video such that
43.8\% of its frames are positive, which is the ratio of positives in the
entire dataset.  Our method significantly outperforms all the existing methods.
We also see our interval proposal idea has a clear positive impact on interval
detection results.  However, when evaluated with the frame classification
metric (first column in Table~\ref{tab:temporalattention_result}), our interval
method does not improve over our frame method.  This is due to some inaccurate
(too coarse) proposals, which may be helped by sampling the level sets more
densely.  We also show an upper bound for the accuracy with perfect interval
hypotheses (see Ours-GT interval), which emphasizes the need to go beyond
frame-wise predictions as we propose.

Fig.~\ref{fig:temporalattention_result} and
Table~\ref{tab:temporalattention_result} show our method performs similarly in
all three train-test settings, meaning it generalizes to both new recorders and
new scenarios.  This is an interesting finding, since it is not obvious \emph{a
priori} that different people exhibit similar motion behavior when they become
engaged, or that those behaviors translate between scenes and activities.  This
is important for applications, as it would be impractical to collect data for
all recorders and scenarios.

The CNN baseline, which learns which video content corresponds to engagement,
does the best of all the baselines.  However,  it is noticeably weaker than our
motion-based approach.  This result surprised us, as we did not expect the
\emph{appearance} of objects in the field of view during engagement intervals to
be consistent enough to learn at all.  However, there are some intra-scenario
visual similarities in a subset of clips: four of the Museum videos are at the
same museum (though the recorders focus on different parts), and five in the
Mall contain long segments in clothing stores (albeit different ones).  Overall
we find the CNN baseline often fails to generate coherent predictions, and it
predicts intervals much shorter than the ground truth.  This suggests that
appearance alone is a weaker signal than motion for the task.

Motion Magnitude (representative
of~\cite{cheatle2004icpr,rallapalli-mobicom2014}) is the next best baseline.
While better than the saliency metrics, its short-term motion and lack of
learning lead to substantially worse results than our approach.
This also reveals that people often move while they engage with objects they want to learn more about.

Finally, despite their popularity in video summarization, Saliency Map
methods~\cite{seo-2009,harel2006nips,itti-2009,ejaz-2013,rahtu-eccv2010,rudoy2013cvpr}
do not predict temporal ego-engagement well.  In fact, they are weaker than the
simpler motion magnitude baseline.  This result accentuates the distinction
between predicting gaze (the common saliency objective) and predicting
first-person engagement.  Clearly, spatial attention does not
directly translate to the task.  While all the Saliency Map
methods (except~\cite{rahtu-eccv2010}) incorporate motion cues, their reliance
on temporally local motion, like flickers, makes them perform no better than
the purely static image methods.

Fig.~\ref{fig:qual} shows example high engagement frames.

\begin{figure}[t]
    \includegraphics[width=1.\linewidth]{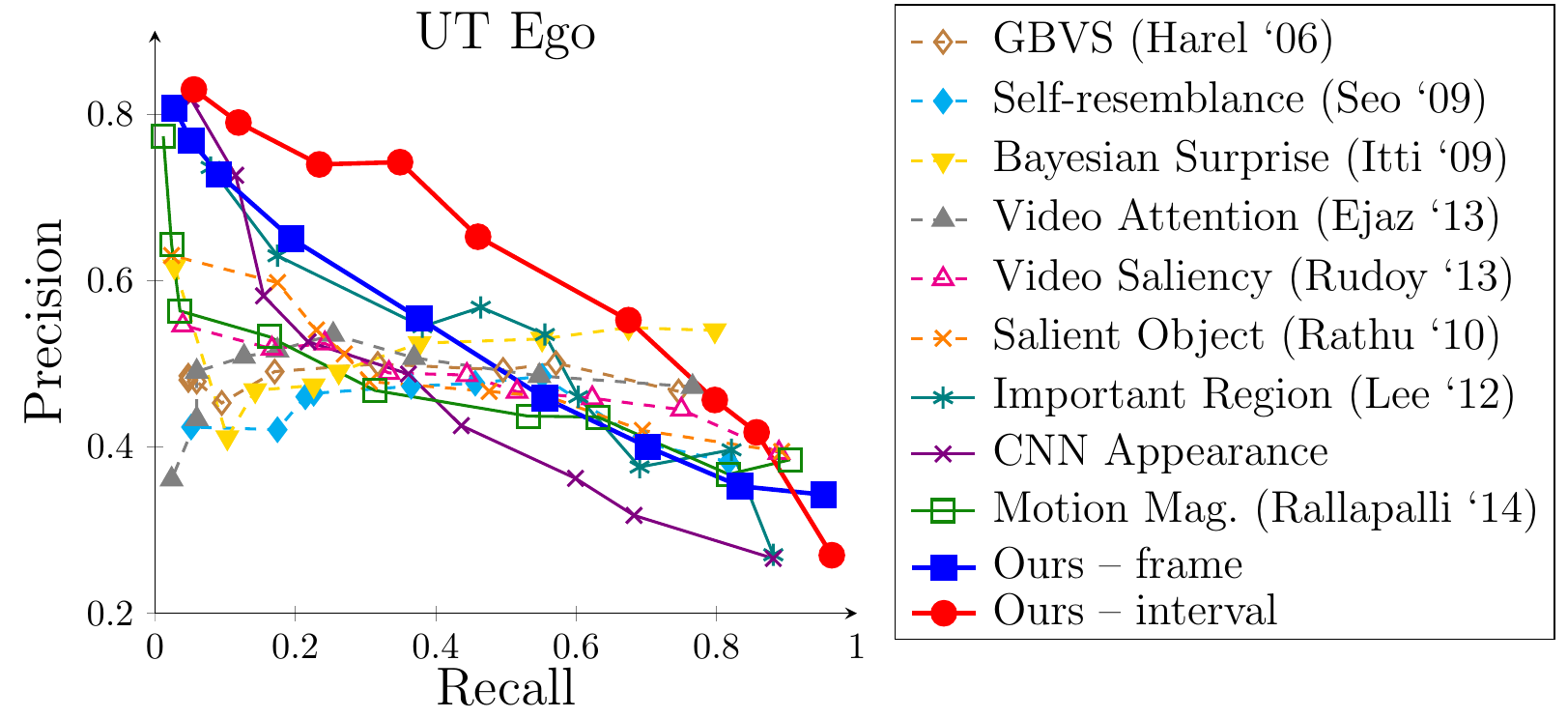}
    \caption{Precision-recall accuracy on UT Ego dataset.}
    \label{fig:utego_result}
\end{figure}

\subsection{UT Egocentric dataset}

Fig.~\ref{fig:utego_result} shows the results on the UT Egocentric dataset.
The outcomes are consistent with those on UT EE above, and again our method
performs the best. Whereas~\cite{lee-cvpr2012} is both trained and tested on UT
Ego, our method does not do any training on the UT Ego data; rather, we use our
model trained on UT EE. This ensures fairness to the baseline (and some
disadvantage to our method).

Our method outperforms the Important Regions~\cite{lee-cvpr2012} method, which
is specifically designed for first-person data.   This result gives further
evidence of our method's cross-scenario generalizability.  Important
Regions~\cite{lee-cvpr2012} does outperform the Saliency Map methods on the
whole, indicating that high-level semantic concepts are useful for detecting
engagement, more so than low-level saliency.  The CNN baseline does poorly,
which reflects that its content-specific nature hinders generalization to a new
data domain.

\subsection{Start point correctness}
\label{sec:startpoint_result}

\begin{figure}[t]
    \includegraphics[width=1.\linewidth]{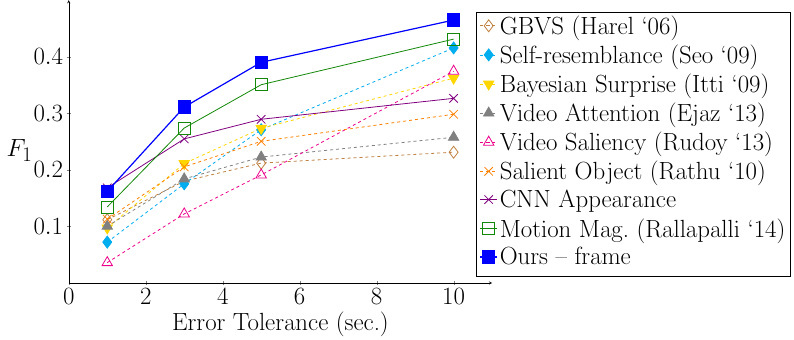}
    \caption{Start-point accuracy on UT EE, measuring how well the onset of an engagement interval is detected in a streaming manner.}
    \label{fig:startpoint_result}
\end{figure}

Finally, Fig.~\ref{fig:startpoint_result} evaluates start point accuracy on UT
EE.  This setting is of interest to applications where it is essential to know
the onset of engagement, but not necessarily its temporal extent.  Here we run
our method in a streaming fashion by using its frame-based predictions, without
the benefit of hindsight on the entire intervals.  To compare the start point
accuracy of different methods, we plot the $F_{1}$ score as a function of error
tolerance window (in seconds) allowed between the predicted and the nearest
ground truth start point. Our method outperforms all other methods under all
error tolerances.  This is evidence that our method has promise for both the
online and offline setting, though we think there remains interesting future
work to best account for streaming data.

The Motion Magnitude baseline is our nearest competitor for this setting.  This
indicates that an abrupt decline in motion is predictive for the transition
between engagement and non-engagement (e.g., as a person slows to examine
something).  However, it remains weaker than our method, and, as we see in the
other results in the main paper, it cannot predict the continuation and
subsequent drop of engagement level.

\section{Conclusion}
We explore engagement detection in first-person video.  By precisely defining
the task and collecting a sizeable dataset, we offer the first systematic study
of this problem.  We introduced a learning-based approach that discovers the
connection between first-person motion and engagement, together with an
interval proposal approach to capture a recorder's long-term motion.  Results
on two datasets show our method consistently outperforms a wide array of
existing methods for visual attention.  Our work provides the foundation for a
new aspect of visual attention research.  In future work, we will examine the
role of external sensors (e.g., audio, gaze trackers, depth) that could assist
in ego-engagement detection when they are available.

\section*{Acknowledgements}

We thank the friends and labmates who assisted us with data collection.  This
research is supported in part by ONR YIP N00014-12-1-0754 and a gift from
Intel.

%
\IEEEpeerreviewmaketitle

\appendices

\section{Annotation Interface}

In this section, we show the interface and instructions for engagement
annotation on Amazon Mechanical Turk.  We include the full instructions and
annotation interface details in order to help reviewers evaluate the care with
which we collected the ground truth annotations.

\begin{figure*}[t]
  	\centering
    \includegraphics[width=1\textwidth,clip,trim={0 4.2cm 0 0}]{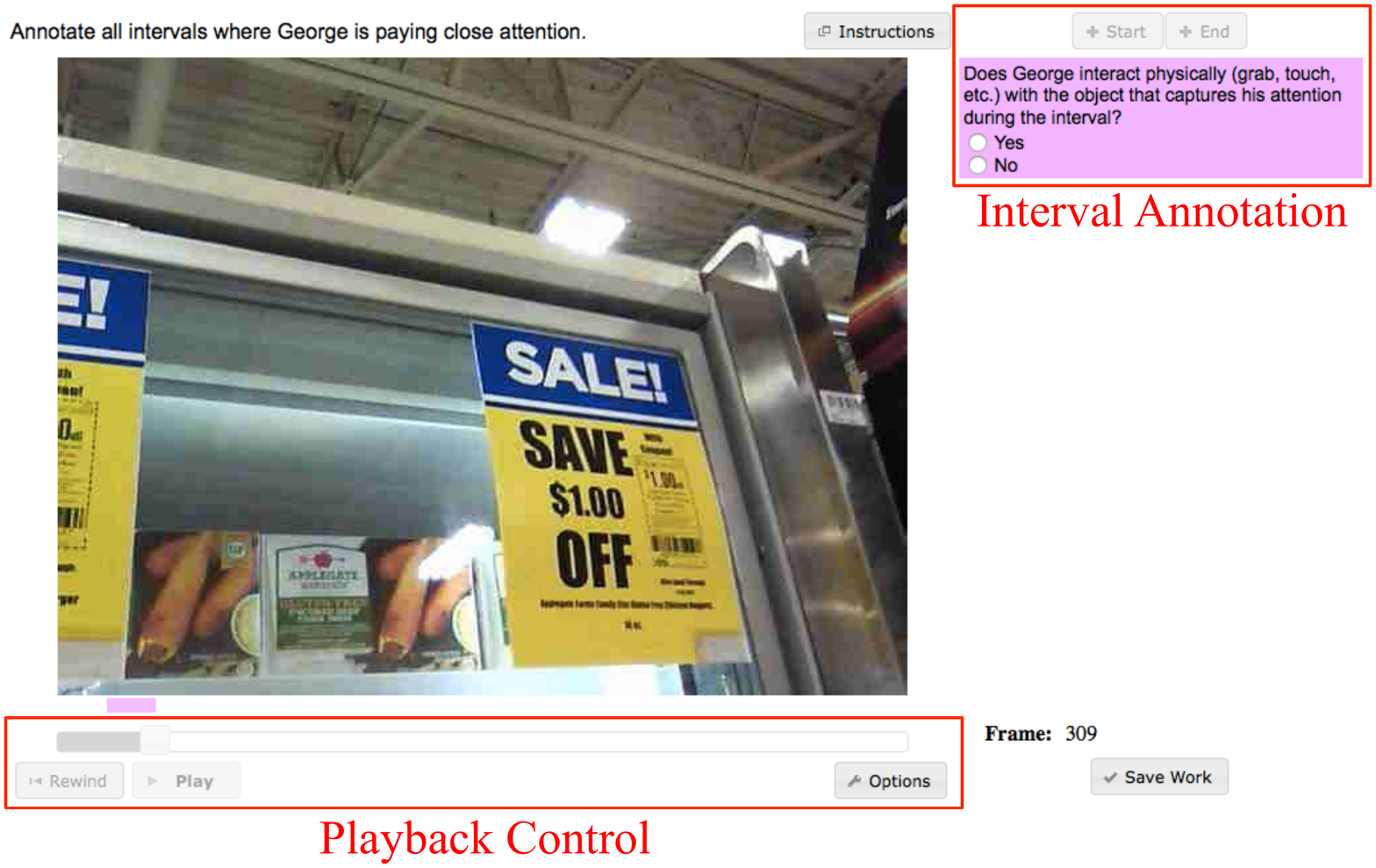}
    \caption{Screen shot of annotation interface.}
\end{figure*}

\subsection{Task Description}\mbox{}

George is wearing a camera on his head. The camera captures video constantly as George goes about his daily life. Because the camera is on his head, when George moves his head to look around, the camera moves too. Basically, it captures the world just as George sees it.

Your job is to watch a video excerpt from George's camera that lasts 1-2 minutes, and determine when \textbf{something in the environment has captured George's attention}. You will first watch the entire video. Then you will go back and use a slider to navigate through the video frames and mark the intervals (start and end points) where he is paying close attention to something. \textbf{Note, the video may have more than one interval where George is paying close attention to something}.

\paragraph{Definition of Attention}\mbox{}

The following instructions will describe what we mean by ``capturing George's attention'' in more detail:
Humans' cognitive process has different levels of attention to the surrounding environment. For example, people pay very little attention to their surroundings when they are walking on a route they are familiar with, but the attention level will rise significantly if there are unusual events (such as a car accident) or something attracts their curiosity (such as a new advertisement on the wall), or if they want to inspect something more closely (such as a product on the shelf when shopping). You are asked to identify these ``high attention intervals'' in the video.

\textbf{In particular, we ask you to identify intervals where George's attention is focussed on an object or a specific location in the scene.}
During these intervals, George is attracted by an object and tries to have a better view/understanding about it intentionally. In general, George may:
\begin{itemize}
	\item Have a closer look at the object
    \item Inspect the object from different views
    \item Stare at the object
\end{itemize}

In some situations, George may even interact physically with the object capturing his attention to gather more information. For example, he may grab the object to have a closer view of it, or he may turn the object to inspect it from different views. To identify these situations, we also ask you to annotate \textbf{whether George touched the object} capturing his attention during the interval.

The following video shows examples of attention interval:
\emph{please refer to the video on our project webpage}.

\paragraph{Important Notes}
	\begin{itemize}
		\item You should watch the entire video (3 minutes) first before doing any annotation. This will give you the context of the activity to know when George is paying close attention.
		\item A video may contain \textbf{multiple or no} intervals where George's attention is captured. You should label each one separately. The intervals are mutually exclusive and should not overlap.
        \item Each interval where George's attention is captured may vary in length. Some could be a couple seconds long, others could be closer to a minute long. The minimum length of each interval is 15 frames (1 second).
        \item You may need to scroll back and forth in the video using our slider interface to determine exactly when the attention starts and stops. Mark the interval as tightly as possible.
		\item After labeling where an attention interval starts and ends, you will mark whether George has physical contact (grab, touch, etc.) with the object during the interval or is just looking at it.
        \item You will also mark your confidence in terms of how strongly George's attention was captured in that interval (Obvious, Fairly clear, Subtle).
    \end{itemize}

\subsection{Interface Introduction}\mbox{}

The following introduction will give you tips on how to best use the tool. Please watch the below video (and/or read the below section) for instructions:
\emph{please refer to the video on our project webpage}.

\paragraph{Getting Started}
\begin{itemize}
	\item Press the \textbf{Play} button to play the video.
	\item After the video finished, press the \textbf{Rewind} button and start annotation.
	\begin{figure}[H]
  		\centering
    	\includegraphics[width=.5\linewidth]{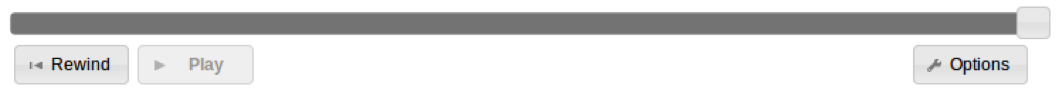}
	\end{figure}
	\item Play the video, \textbf{Pause} the video when you reach the frame at the beginning of high attention interval.
	\item Click the \textbf{Start} button to mark the ``Start'' of the interval.
	\begin{figure}[H]
  		\centering
    	\includegraphics[width=.5\linewidth]{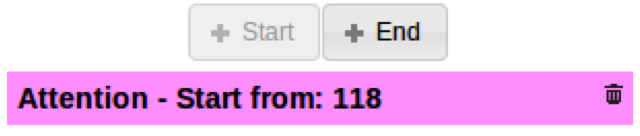}
	\end{figure}
	\item On the right, directly below the Start button, you will find a colorful box showing the frame number corresponding to the `Start' of the interval.

	\item Similarly, click the \textbf{End} button to mark the ``End'' of the interval.

	\item After you mark the end of the interval, you will be asked whether George contact (grabbing, touching, etc.) the object that captures his attention.
	
	\item Next, you will be asked about how obvious is the attention interval. Specify whether the interval is \textbf{Obvious, Fairly clear, Subtle}.
	\begin{figure}[H]
  		\centering
    	\includegraphics[width=.5\linewidth]{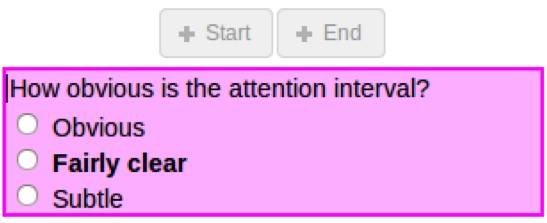}
	\end{figure}

	\item Finally, you will be asked to describe what attracts George's attention. Type in what attracts George's attention (object, scene, event, etc.) and \textbf{Submit} the interval.
	\begin{figure}[H]
  		\centering
    	\includegraphics[width=.5\linewidth]{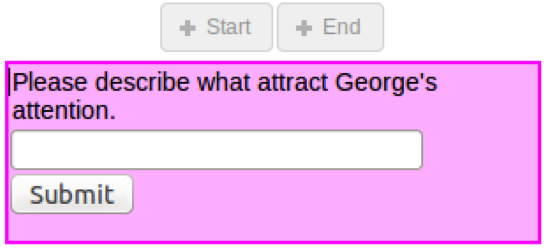}
	\end{figure}
  
	\item When you are ready to submit your work, rewind the video and watch it through one more time. Do the ``Start'' and ``End'' you specified cover the complete high attention interval? After you have checked your work, press the \textbf{Submit HIT} button. We will pay you as soon as possible.

	\item Do \textbf{not} reload or close the page before redirected to next hit. This may cause submission failure.
\end{itemize}

\paragraph{How We Accept Your Work}\mbox{}

We will hand review your work and we will only accept high quality work. Your annotations are not compared against other workers.

\paragraph{Keyboard Shortcuts}\mbox{}

These keyboard shortcuts are available for your convenience:
\begin{itemize}
    \item \textbf{t} toggles play/pause on the video
    \item \textbf{r} rewinds the video to the start
    \item \textbf{d} jump the video forward a bit
    \item \textbf{f} jump the video backward a bit
    \item \textbf{v} step the video forward a tiny bit
    \item \textbf{c} step the video backward a tiny bit
\end{itemize}

\ifCLASSOPTIONcaptionsoff
  \newpage
\fi



%

\bibliographystyle{IEEEtran}
\bibliography{kgrefs-attention,egocentricengagement}

\end{document}